%% file: main.tex
\documentclass{article}
\input{macros}
\usepackage{amsmath}

\usepackage[preprint,nonatbib]{neurips_2020}

\usepackage[utf8]{inputenc} 
\usepackage[T1]{fontenc}    
\usepackage{hyperref}       
\usepackage{url}            
\usepackage{booktabs}       
\usepackage{amsfonts}       
\usepackage{nicefrac}       
\usepackage{microtype}      

\title{Solving Math Word Problems by Combining Language Models With Symbolic Solvers} 

\author{%
  Joy He-Yueya\\
  Stanford University\\
  \texttt{heyueya@stanford.edu} \\
  \And
  Gabriel Poesia \\
  Stanford University\\
  \texttt{poesia@cs.stanford.edu}
  \And
  Rose E. Wang \\
  Stanford University\\
  \texttt{rewang@stanford.edu}
  \And
  Noah D. Goodman \\
  Stanford University\\
  \texttt{ngoodman@stanford.edu}
}

\begin{document}

\maketitle

\begin{abstract}
Automatically generating high-quality step-by-step solutions to math word problems has many applications in education. Recently, combining large language models (LLMs) with external tools to perform complex reasoning and calculation has emerged as a promising direction for solving math word problems, but prior approaches such as Program-Aided Language model (\pal) are biased towards simple procedural problems and less effective for problems that require declarative reasoning. We propose an approach that combines an LLM that can incrementally formalize word problems as a set of variables and equations with an external symbolic solver that can solve the equations. Our approach achieves comparable accuracy to the original $\pal$ on the $\gsm$ benchmark of math word problems and outperforms $\pal$ by an absolute $20\%$ on $\algebra$, a new dataset of more challenging word problems extracted from Algebra textbooks. Our work highlights the benefits of using declarative and incremental representations when interfacing with an external tool for solving complex math word problems. Our data and prompts are publicly available at \href{https://github.com/joyheyueya/declarative-math-word-problem}{https://github.com/joyheyueya/declarative-math-word-problem}.
\end{abstract}

\input{introduction}
\input{related_work}
\input{our_approach}
\input{experimental_setup}
\input{results}
\input{conclusion}

\bibliography{main}
\bibliographystyle{abbrv}

\end{document}

%% file: macros.tex
\newcommand{\algebra}{\textsc{Algebra}}
\newcommand{\gsm}{\textsc{GSM8k}}
\newcommand{\pal}{\textsc{PAL}}
\newcommand{\chain}{\textsc{CoT}}
\newcommand{\declarative}{\textsc{Declarative}}
\newcommand{\onestep}{\textsc{One-step Declarative}}
\newcommand{\sympy}{\textrm{SymPy}}
\usepackage{subcaption}

\usepackage[ruled,vlined]{algorithm2e}

\usepackage{url}

\usepackage{dsfont}

\usepackage{enumitem}
\usepackage{hyperref}
\usepackage{listings}
\usepackage{url}
\usepackage{graphicx}
\usepackage{xcolor}

%% file: introduction.tex
\section{Introduction}

\begin{figure}
    \centering    \includegraphics[width=0.5\columnwidth, height=0.454\columnwidth]{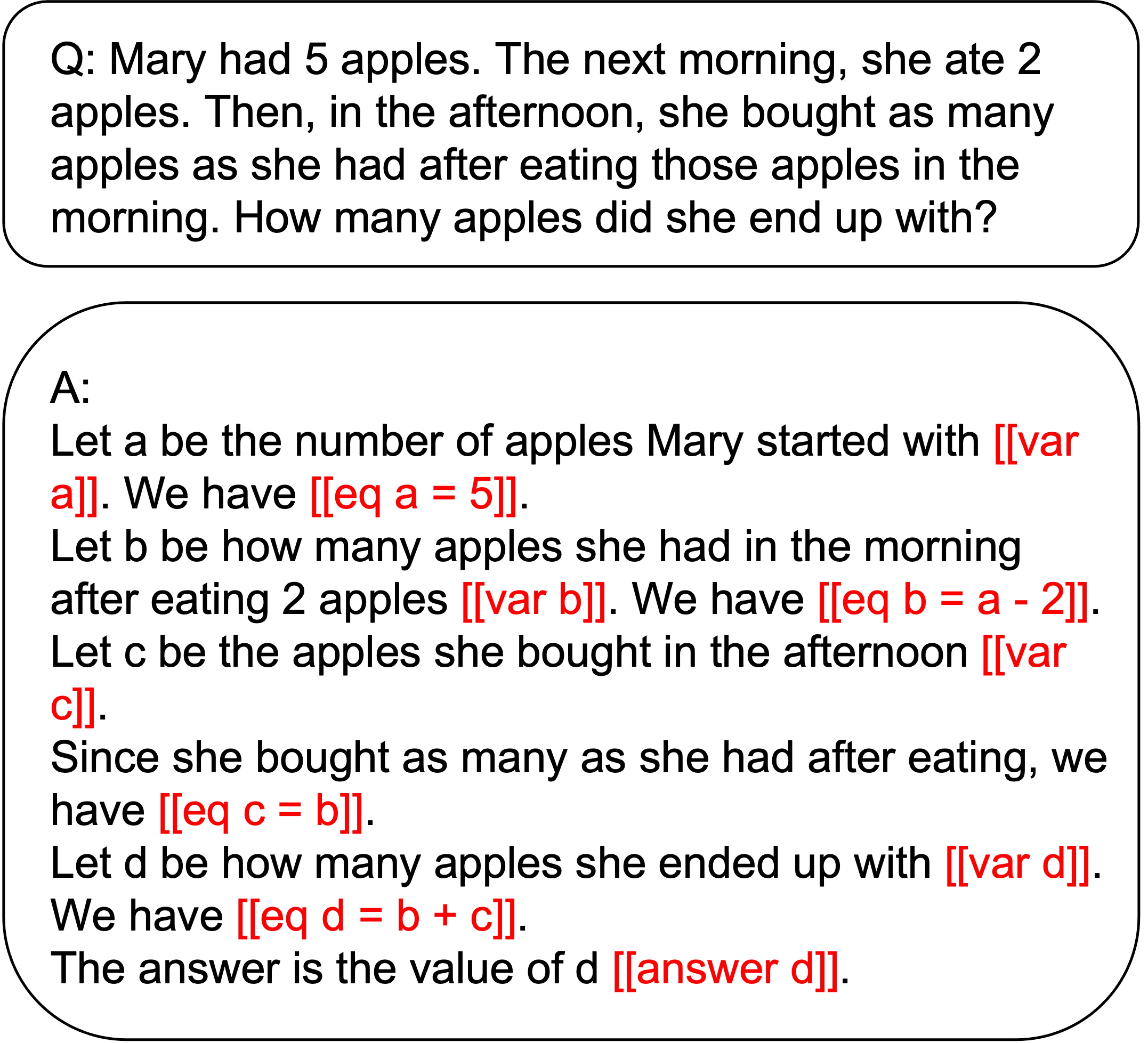}
    \caption{An example of a math word problem and its solution from the $\declarative$ prompt. Variables and equations are in \textcolor{red}{red}.}   \label{fig:declarative_prompt_example}
\end{figure}

Learning to solve mathematical word problems (see an example in Figure \ref{fig:declarative_prompt_example}) is an important skill but can be challenging for students. \cite{cummins1991children, pongsakdi2020makes}. A tool that can automatically generate step-by-step solutions to such problems has the potential to provide personalized support for students working through word problems \cite{ritter2007cognitive, del2022intelligent} and help educators with curriculum development \cite{polozov2015personalized}.

Using few-shot prompting over large language models (LLMs) has recently emerged as a promising approach for solving math word problems \cite{wei2022chain, zhou2022least, gao2022pal}. The chain-of-thought (\chain) \cite{wei2022chain} prompting method presents explicit intermediate reasoning steps to the LLM to further enhance its reasoning capability. However, LLMs often struggle with performing arithmetic operations \cite{hendrycks2021measuring, lewkowycz2022solving, wei2022chain}. To address this, \cite{wei2022chain} uses an external calculator to evaluate the arithmetic operations in the generated reasoning steps. Program-Aided Language model (\pal) \cite{gao2022pal} extends this idea by generating Python programs as
reasoning steps, offloading all calculations to a Python interpreter.
Although programs offer a direct representation of \emph{procedures},
they require special devices to represent more abstract mathematical \emph{declarations}. For example, a statement like $a = b + 1$ can be directly
interpreted as a variable assignment in Python if $b$ is known, but not if $b$ is unknown. Nonetheless, the equation remains a valid mathematical expression even when $b$ is unknown,
suggesting that we instead want to allow models to perform mathematical \emph{declarations} beyond those that yield a procedure (for a full example, see the problem in Figure \ref{fig:declarative_vs_procedural}).

In this work, we present an approach that combines an LLM, which can incrementally formalize word problems as a set of variables and equations, with an external symbolic solver that can solve the equations. Our approach achieves comparable performance to the original $\pal$ on the $\gsm$ \cite{cobbe2021training} benchmark of math word problems. To evaluate current approaches on more challenging word problems, we introduce $\algebra$, a dataset of $222$ word problems collected from open access Algebra textbooks. We show that our approach outperforms $\pal$ by an absolute $20\%$ on $\algebra$. Our work highlights the effectiveness of incrementally generating declarative formalizations when interfacing with an external tool for solving complex math word problems. 

%% file: related_work.tex
\section{Related work}
Recent studies have explored the use of few-shot prompting over LLMs for solving math word problems \cite{wei2022chain, zhou2022least, gao2022pal}. The chain-of-thought \cite{wei2022chain} prompting method presents explicit intermediate reasoning steps to the LLM to improve its reasoning capability. Since LLMs often make arithmetic errors \cite{hendrycks2021measuring, lewkowycz2022solving, wei2022chain}, several prior works \cite{wei2022chain, chowdhery2022palm} have experimented with using an external calculator to carry out the operations generated by LLMs. This generally improves final performance by less than $5\%$ on \gsm. Program-Aided Language model \cite{gao2022pal} extends to more complex arithmetic by generating Python programs as reasoning steps and using a Python interpreter to perform the calculations. However, generating Python programs carries a strong bias toward procedural calculations and does not work well for word problems that do not have a straightforward procedural solution.

%% file: our_approach.tex
\section{Our Approach: Equipping an LLM With an External Symbolic Solver}

Our approach for solving a math word problem consists of two steps: \textit{ (1) declarative and incremental formalization using an LLM} and \textit{(2) solving equations using a symbolic solver}.

\subsection{Declarative and incremental formalization using an LLM}

To solve a math word problem, we first use an LLM to formalize the problem as a set of variables and equations. Recently, using few-shot prompting over LLMs has emerged as an effective approach for natural language understanding and decomposition.

\subsubsection{Few-shot prompting}
Few-shot prompting is a technique that uses LLMs to solve a task by providing the LLMs with a few demonstrations of the task as part of the input at inference time \cite{brown2020language}. In this technique, the demonstrations (i.e., examples of input-output pairs) are concatenated into a prompt, which is passed to the model along with the new input to generate an output. Formally, a set of $k$ input-output examples $\{(x_i, y_i)\}_{i=1}^{k}$ are concatenated in a prompt $p \equiv (x_1, y_2) || (x_1, y_2) || ... || (x_k, y_k)$ where $||$ denotes the concatenation of examples. At inference time, $p || x_{test}$ is passed to the model where $x_{test}$ denotes a new input instance, and the model attempts to complete $p || x_{test}$ by generating the output $y_{test}$.

\subsubsection{Crafting the \declarative ~prompt}

To formalize word problems using few-shot prompting, we introduce the $\declarative$ prompt $p \equiv (x_1, y_2) || (x_1, y_2) || ... || (x_k, y_k)$ where $x_i$ is the word problem in natural language, and $y_i$ is the step-by-step solution to $x_i$. In the $\declarative$ prompt, $y_i$ consists of interleaved natural language statements and formal variable or equation declarations in double-square brackets. Our approach aims to generate solutions that formalize word problems based on a set of principles listed in Table \ref{tab:principles}. Figure \ref{fig:declarative_prompt_example} shows an example used in the $\declarative$ prompt that we created according to these principles. The full prompt is publicly available at \href{https://github.com/joyheyueya/declarative-math-word-problem}{https://github.com/joyheyueya/declarative-math-word-problem}. To solve a new word problem, $x_{test}$, we append it to $p$ and pass $p || x_{test}$ to an LLM, which generates $y_{test}$ as the solution for $x_{test}$.

\begin{table}[ht]
  \centering 
  \begin{tabular}{l}
    \toprule
         Principles for solutions\\
    \midrule
    1. Each sentence in the solution either introduces a new variable or states a new equation. \\
    2. The last sentence gives the goal: which variable will contain the answer to the problem. \\ 
    3. Each equation only uses previously introduced variables. \\
    4. Each quantity is only named by one variable. \\
    5. The solution uses all the numbers in the question.\\
    \bottomrule
  \end{tabular}
  \vspace{5mm} 
  \caption{A list of principles we would like the solutions to satisfy.}
  \label{tab:principles}
\end{table}

\subsection{Solving equations using a symbolic solver}

The step-by-step solution generated by the LLM using the $\declarative$ prompt includes the list of variables and equations that describe the word problem but does not provide the final answer (see Figure \ref{fig:declarative_prompt_example}). Instead of relying on the LLM to solve the equations directly, we pass the equations to an external symbolic solver to do the calculation. In this work, we use \sympy ~\cite{meurer2017sympy}, a Python library for symbolic computation, to algebraically solve a system of equations extracted from the solution generated by the LLM.

%% file: experimental_setup.tex
\section{Experimental Setup}
\subsection{Datasets}

We evaluate our approach on two math word problem datasets: \gsm ~\cite{cobbe2021training} and a new dataset called $\algebra$ \footnote{The $\algebra$ dataset is publically available at \href{https://github.com/joyheyueya/declarative-math-word-problem}{https://github.com/joyheyueya/declarative-math-word-problem}.}. We use the $\gsm$ test set, which contains $1319$ math word problems at grade-school level. To evaluate our approach on more challenging problems, we curated \algebra, which consists of $222$ word problems from two open-access Algebra textbooks: Basic Algebra with Applications (\cite{zaigralin_2018}; released under the Creative Commons Attribution-ShareAlike license) and Elementary Algebra 2e (\cite{marecek_anthony-smith_mathis_2020}; released under the Creative Commons Attribution license).
The resulting dataset includes word problems covering all topics leading up to System of Equations, with the exception of problems related to geometry, graphing, or inequalities.

\subsection{Baselines and variants of the \declarative ~prompting}

\begin{figure}
     \centering
     \begin{subfigure}[b]{0.48\textwidth}
         \centering
         \includegraphics[width=\textwidth]{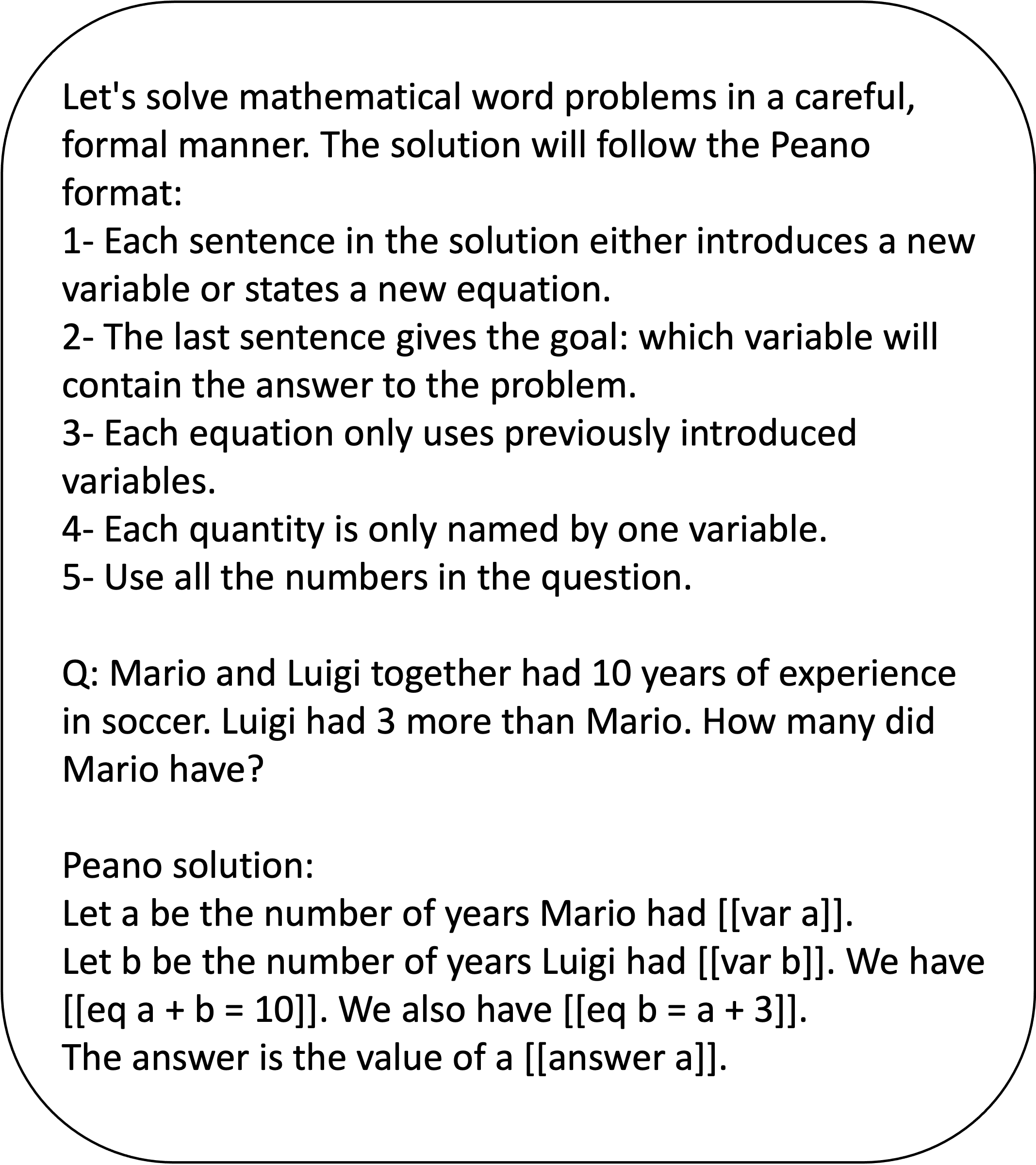}
         \caption{Adding principles to the beginning of the \declarative ~prompt.
         \\\hspace{\textwidth}
         }
         \label{fig:declarative_principle_sympy}
     \end{subfigure}
     \hfill
     \begin{subfigure}[b]{0.48\textwidth}
         \centering
         \includegraphics[width=\textwidth]{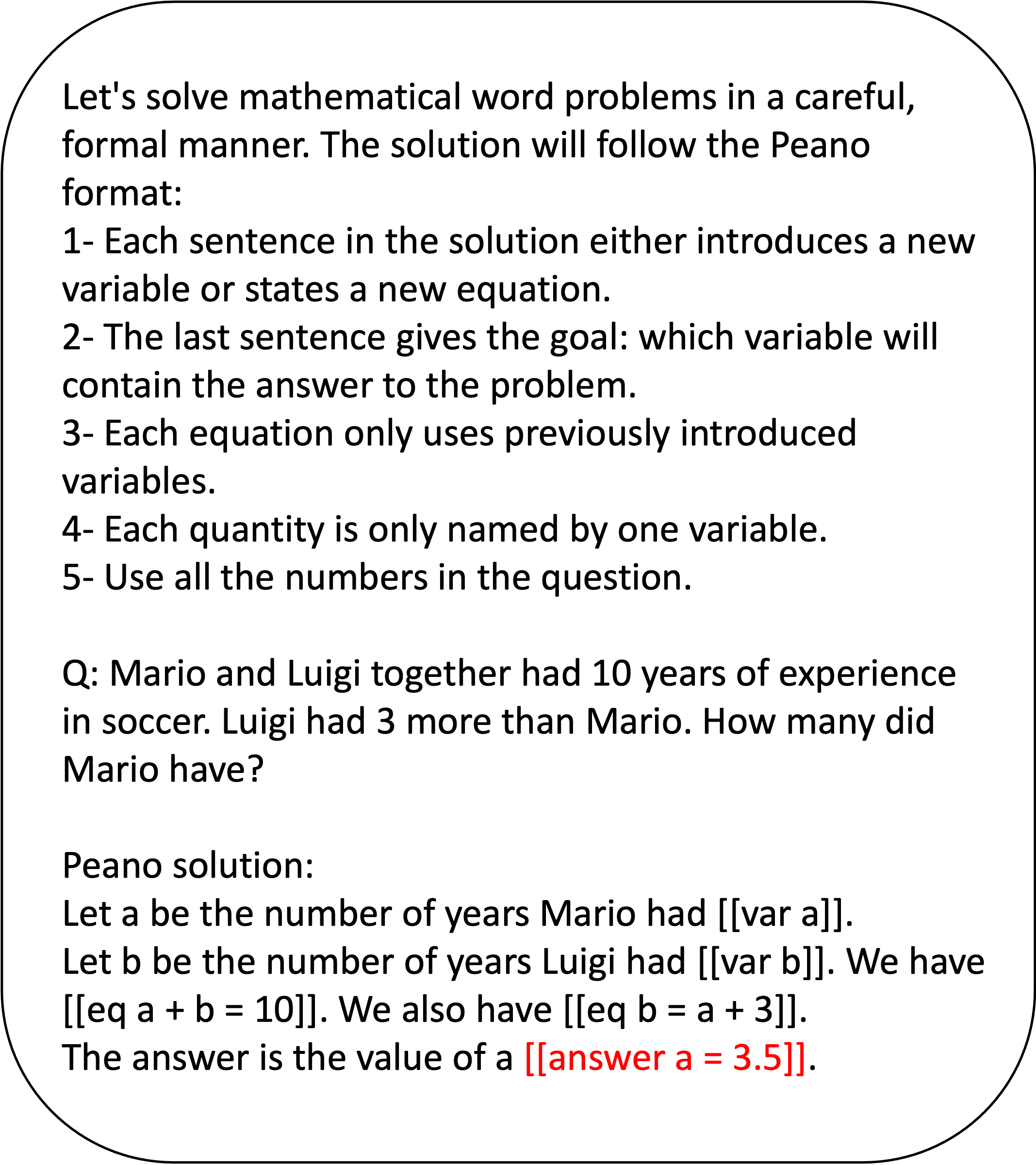}
         \caption{Adding principles to the beginning of the \declarative ~prompt and calculating the final answer. The final answer is in \textcolor{red}{red}.}
         \label{fig:declarative_principle}
     \end{subfigure}
        \caption{The difference between ``$\declarative_{\textrm{3-shot}} + \textrm{principles} + \sympy$'' and ``$\declarative_{\textrm{3-shot}} + \textrm{principles}$'' is that ``$\declarative_{\textrm{3-shot}} + \textrm{principles} + \sympy$'' passes the equations to $\sympy$ to solve, but ``$\declarative_{\textrm{3-shot}} + \textrm{principles}$'' asks the LLM to solve the equations directly.}
        \label{fig:add_principle}
\end{figure}

\begin{figure}
    \centering    \includegraphics[width=0.5\columnwidth]{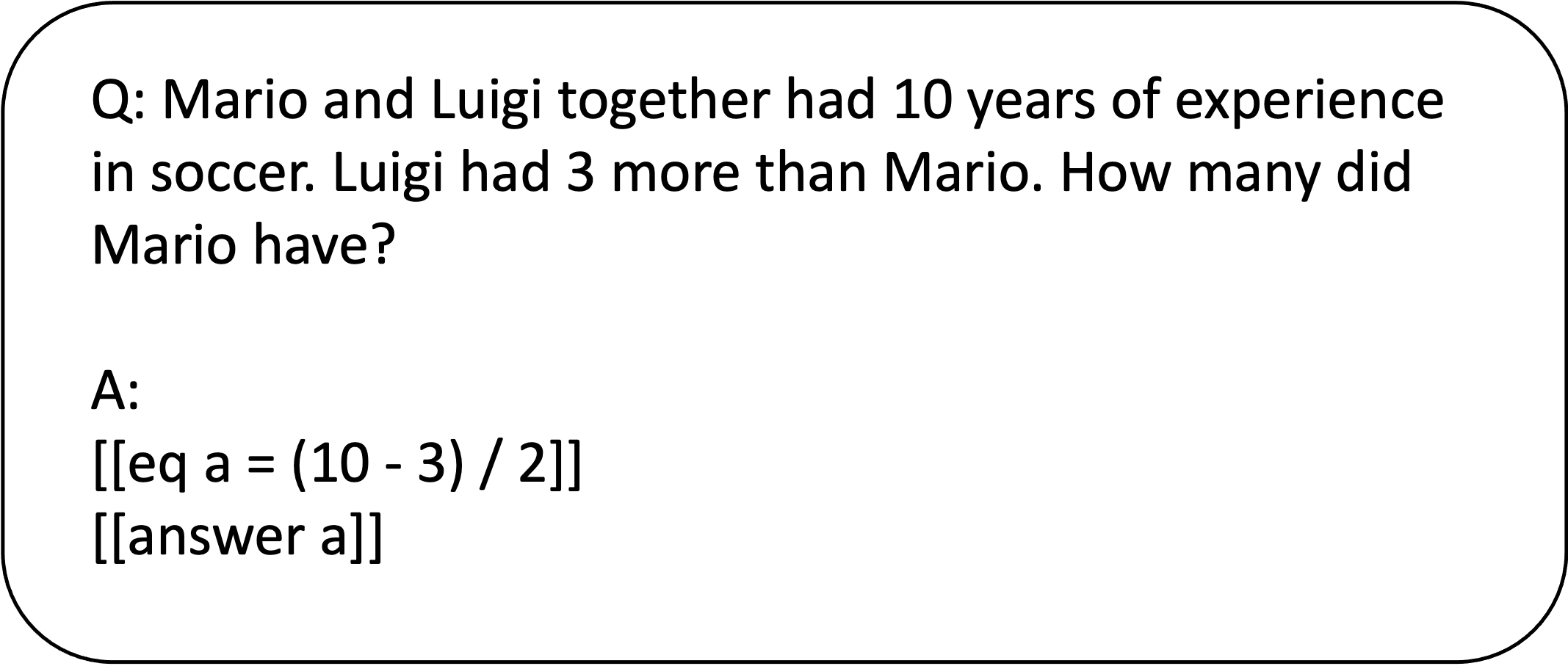}
    \caption{An example of formalizing a math word problem in a single equation.} 
    \label{fig:one_step}
\end{figure}

We consider three methods: chain-of-thought (\chain) prompting \cite{wei2022chain}, Program-Aided Language model (\pal) \cite{gao2022pal}, and our \declarative ~prompting combined with \sympy ~($\declarative + \sympy$). We created two different prompts for each prompting method. The first prompt (8-shot) uses the same set of eight examples used in prior work \cite{wei2022chain}. The second prompt (3-shot) uses three examples that we designed to help illustrate step-by-step and declarative thinking and the formalization format we expect.

For our \declarative ~prompting method, we experimented with three variants. 

\begin{enumerate}
    \item $\declarative_{\textrm{3-shot}} + \textrm{principles} + \sympy$: adding the list of principles in Table \ref{tab:principles} at the beginning of the prompt (see an example in Figure \ref{fig:declarative_principle_sympy}).
    \item $\declarative_{\textrm{3-shot}} + \textrm{principles}$: using the LLM to directly calculate the value of the goal variable (see an example in Figure \ref{fig:declarative_principle}).
    \item $\onestep_{\textrm{3-shot}} + \sympy$: formalizing the word problem in a single step instead of incrementally (see an example in Figure \ref{fig:one_step}).
\end{enumerate}

All the prompts used in this work are publicly available at \href{https://github.com/joyheyueya/declarative-math-word-problem}{https://github.com/joyheyueya/declarative-math-word-problem}.

We use Codex (code-davinci-002) \cite{chen2021evaluating} as the LLM for all methods. We use top-1 decoding and a temperature of 0. We set \texttt{max\_tokens} to be $600$.

%% file: results.tex
\section{Results}
\label{results}

\subsection{Results on \gsm ~and \algebra}

\begin{table}[ht]
  \label{sample-table}
  \centering
  \begin{tabular}{lll}
    \toprule
         & \gsm & \algebra \\
    \midrule
    $\chain_{\textrm{8-shot (original)}}$ & $62.5 \pm 0.16$ & $45.3 \pm 0.56$   \\
    $\chain_{\textrm{3-shot (ours)}}$ & $58.9 \pm 0.16$ & $47.9 \pm 1.18$   \\
    $\pal_{\textrm{8-shot (original)}}$ & $70.2 \pm 0.25$ & $51.7 \pm 0.21$     \\
    $\pal_{\textrm{3-shot (ours)}}$ & $\textbf{73.3} \pm 0.13$ & $56.2 \pm 0.21$     \\
    $\declarative_{\textrm{8-shot}} + \sympy$ & $64.7$ & -     \\
    $\declarative_{\textrm{3-shot}} + \sympy$ & $66.0 \pm 0.33$ & -     \\
    $\declarative_{\textrm{3-shot}} + \textrm{principles} + \sympy$ & $69.4 \pm 0.65$ & $\textbf{76.3} \pm 0.93$    \\
    $\declarative_{\textrm{3-shot}} + \textrm{principles}$ & $22.4 \pm 0.27$ & -    \\
    $\onestep_{\textrm{3-shot}} + \sympy$ & $57.5 \pm 0.06$ & -    \\
    \bottomrule
  \end{tabular}
  \vspace{5mm} 
  \caption{
  Problem solve rate (\%) on \gsm ~and \algebra. We report the average and standard deviation across three runs. The highest number on each dataset is in \textbf{bold}. For \chain ~and \pal, we ran both the 8-shot prompt used in the original papers and the 3-shot prompt we created.
  }
  \label{tab:main_results}  
\end{table}

\begin{figure}[ht]
    \centering    \includegraphics[width=\columnwidth]{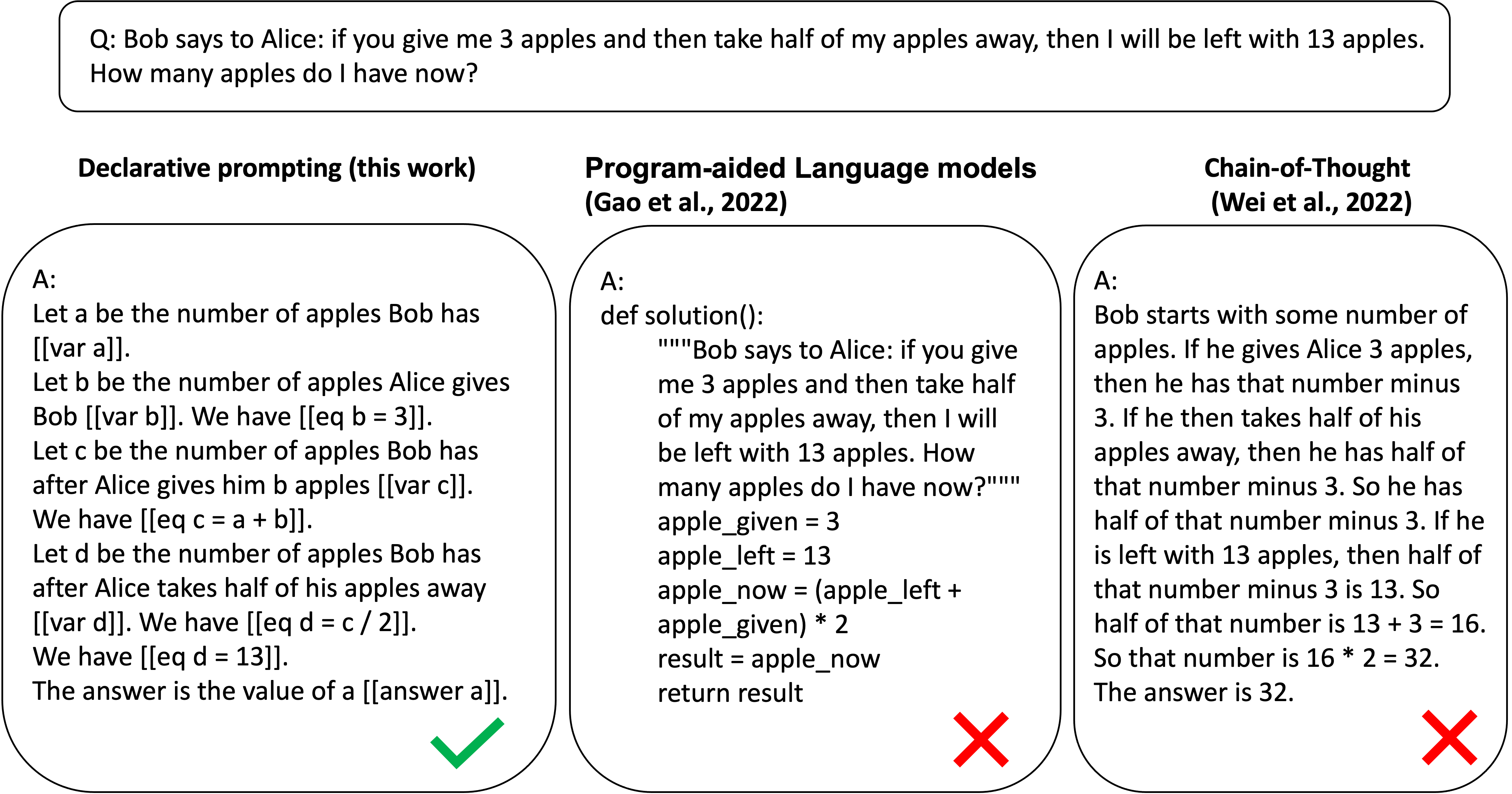}
    \caption{Declarative solutions are typically more intuitive to write than procedural solutions for challenging algebra word problems. \pal ~and \chain ~try to generate procedural solutions that describe a set of plans for achieving the goal, which are incorrect in this case. The \declarative ~prompting generates a correct solution that describes the properties of the goal, which is generally more appropriate for hard problems with no obvious procedural solutions. } 
    \label{fig:declarative_vs_procedural}
\end{figure}

On $\gsm$ (Table \ref{tab:main_results}), our 3-shot prompt leads to a better performance than the original 8-shot prompt for \pal ~and \declarative. \pal ~outperforms \declarative ~across both sets of comparable examples, but using our \declarative ~prompting method with the 3-shot prompt ($\declarative_{\textrm{3-shot}} + \textrm{principles} + \sympy$) gives a performance equivalent to the original \pal ~($\pal_{\textrm{8-shot (original)}}$).

Interestingly, prepending the list of principles to the \declarative ~prompt ($\declarative_{\textrm{3-shot}} + \textrm{principles} + \sympy$) leads to a better performance on \gsm ~than $\declarative_{\textrm{3-shot}} + \sympy$. Asking the LLM to solve the equations directly leads to a dramatic drop in accuracy (from $69.4\%$ to $22.4\%$), which highlights the benefit of using an external solver. Additionally, our \declarative ~prompting benefits from incremental formalization, as shown by the performance gap between the incremental version ($\declarative_{\textrm{3-shot}} + \textrm{principles} + \sympy$) and the non-incremental variant ($\onestep_{\textrm{3-shot}} + \sympy$). 

On \algebra ~(Table \ref{tab:main_results}), our approach ($\declarative_{\textrm{3-shot}} + \textrm{principles} + \sympy$) achieves the highest accuracy among all methods, outperforming $\pal$ by an absolute $20\%$. The accuracy of the original $\chain$ drops from $62.5\%$ on $\gsm$ to $45.3\%$ on \algebra, which demonstrates that problems in \algebra ~are generally harder than those in \gsm. The main reason that the \declarative ~prompting method works better than \chain ~and \pal ~on \algebra ~is that it is less intuitive to generate procedural solutions to Algebra problems that require declarative reasoning (see an example in Figure \ref{fig:declarative_vs_procedural}). Although our 3-shot prompt improves the performance of $\chain$ and $\pal$ on $\algebra$ compared to the original 8-shot prompt, our $\declarative$ method is still much more effective than $\chain$ and $\pal$.

%% file: conclusion.tex
\section{Conclusion}

We present an approach for automatically generating step-by-step solutions to math word problems by equipping an LLM with an external symbolic solver. Our approach uses an LLM to incrementally formalize word problems as variables and equations and avoids arithmetic errors by using an external symbolic solver that can solve the equations. Our approach achieves comparable accuracy to the original $\pal$ on $\gsm$ and improves over $\pal$ by an absolute $20\%$ on a new dataset consisting of harder word problems from Algebra textbooks.

We demonstrate the effectiveness of using declarative formalization when interfacing with an external tool for solving complex math word problems. Additionally, encouraging incremental formalization is beneficial, especially when using declarative representations. Our approach is particularly useful for math education since many advanced math problems can be divided into separate conceptual pieces, with one piece being declarative and the other involving procedural knowledge.